%% file: artigoPropor.tex
\documentclass[runningheads]{llncs}
\usepackage{makeidx}  

\usepackage{verbatim}
\usepackage{graphicx}
\usepackage{url}
\usepackage{multirow}
\usepackage{threeparttable}
\usepackage{epstopdf}

\input{macros}

\begin{document}
\frontmatter          
\pagestyle{headings}  
%

%
\mainmatter              

\title{\thisTitleA}

\author{Juliana~P.~C.~Pirovani\inst{1}~\textsuperscript{https://orcid.org/0000-0002-4157-6503} \and
Elias~de~Oliveira\inst{1} \and 
Eric~Laporte\inst{2}~\textsuperscript{https://orcid.org/0000-0002-0984-0781}}

\authorrunning{Pirovani et al.}

\institute{Universidade Federal do Esp\'irito Santo - UFES \\ Av. Fernando Ferrari, 514, 29075-910 Vit\'oria, ES, Brazil \\
\email{jupcampos@gmail.com, elias@lcad.inf.ufes.br}\and
Universit\'e Paris-Est, \\ LIGM, UPEM/CNRS/ENPC/ESIEE, Champs-sur-Marne, 77420, France\\
\email{eric.laporte@univ-paris-est.fr}
}

\maketitle

\begin{abstract}
\input{abstract-en.tex}

\end{abstract}

\section{Introduction}
\label{intro}
\input{introduction-en.tex}


\section{The Methodology}
\label{theMethodology}
\input{methodology-en.tex}

\section{Results and Discussion}
\label{results}
\input{results-en.tex}

\section{Conclusions}
\label{theconclusion}
\input{conclusion-en.tex}

\bibliography{bibliografia-Juliana}
\bibliographystyle{splncs04}

\end{document}

%% file: macros.tex
\newcommand{\thisTitleA} {Concordance Comparison as a Means of Assembling Local Grammars}

\newcommand{\thisApproach} {LG}

\newcommand{\Precision}{79.75}
\newcommand{\Recall}{74.18}
\newcommand{\Fmeasure}{76.86}

\newcommand{\PrecisionPG}{81}
\newcommand{\RecallPG}{60}
\newcommand{\FmeasurePG}{69}
\newcommand{\GI}{$G_1$}
\newcommand{\GII}{$G_2$}
\newcommand{\CI}{$C_1$}
\newcommand{\CII}{$C_2$}
\newcommand{\CIxCII}{$C_1 \times C_2$}

%% file: abstract-en.tex
%
%

Named Entity Recognition for person names is an important but non-trivial task in information
extraction. This article uses
a tool that compares the concordances obtained from two local grammars (LG) and 
highlights the differences. We used the results as an aid to select the best of a set of LGs.
By analyzing the comparisons, we observed relationships of inclusion, intersection and disjunction within each pair of LGs,
which helped us to assemble those that yielded the best results.
This approach was used in a case study on extraction of person names from 
texts written in Portuguese.  We applied the enhanced grammar to the Gold Collection of the Second HAREM. The F-Measure obtained was {\Fmeasure}, representing a gain
of 6 points in relation to the state-of-the-art for Portuguese.

%% file: introduction-en.tex
%
%

Named Entity Recognition (NER) involves automatically identifying names of entities such as persons, places and organizations. Person names are a fundamental source of information. Many applications seek information on individuals and their relationships,
e.g.\ in the context of social networks. However, extracting this type of Named Entity (NE) is challenging:
person names are an open word class, which includes many words and grows every day \cite{manning1999}. 

``\textit{A good portion of NER research is devoted to the study of English, due to its significance
as a dominant language that is used internationally}'' \cite[page 470]{shaalan2014:cl}. An influential impetus to the development of
systems for this purpose in Portuguese came with the HAREM \cite{santos2007,mota2008} events, a joint assessment of the area organized by
Linguateca \cite{linguateca2018:site}. The annotated corpora used in the first and second
HAREM, known as the Golden Collection (GC), serve as a reference for recent works on Portuguese NER.

The main approaches used to develop NER systems involve (i) machine learning, whereby systems learn to identify and classify NEs from a training
corpus, (ii) the linguistic approach, which involves manual description of rules in which NEs can appear, and (iii) a hybrid 
approach that combines both previous methods. 

``\textit{Local grammars (LG) are finite-state grammars or finite-state automata that represent sets of utterances of a natural language}'' 
\cite[page 1]{gross1999:scm}. They were introduced by Maurice Gross \cite{gross1997:fslp} and serve
as a way to group phrases with common characteristics (usually syntactic or semantic). Describing rules in the form of LGs for the construction of Information
Extraction (IE) systems requires human expertise and training in linguistics; little computational aid for this task is available. 

A method for constructing LGs around a keyword or semantic unit is presented by \cite{gross1999:scm}. LGs for extracting person names from Portuguese texts were
presented in \cite{baptista1998} and \cite{pirovani2015:cob}. In the Second HAREM \cite{mota2008}, the Rembrandt system, which uses grammar rules and Wikipedia as sources of knowledge \cite{cardoso2008}, ranked best for the `person' category.
A comparison between four tools to recognize NEs in Portuguese texts
\cite{amaral2014:lrec} suggested that the rule-based approach is the most effective for person names.
Recently, LGs have been successfully integrated in a hybrid approach to Portuguese NER \cite{pirovani2017:isda}.

This paper describes how to use the Unitex concordance comparison tool \cite{unitex2018:site} as an aid to constructing
an LG. Our point of departure was a set of LGs to identify person names in Portuguese texts.
By comparing concordances obtained from them,
we found some relationships between them in terms of set theory. Taking into account these relationships,
we picked the best LGs and combined them in order to achieve better performance.

This article is organized as follows. Section 2 presents the methodology used in this work. The results of the study are presented in Section 3,
and Section 4 presents conclusions and avenues for future research.

%% file: methodology-en.tex
%
%

The input to our experiment was a repository of small LGs to recognize person names.
Some were obtained from the literature (e.g. those presented in \cite{baptista1998}) and we created others.

All of these LGs were created and processed with Unitex \cite{unitex2018:site}, an open-source system initially developed
at University of Paris-Est Marne-La-Vallée in France. A local grammar is represented as a set of one or more graphs referred 
to as Local Grammar Graphs (LGG). Unitex allows for creating LGGs, preprocessing texts, applying dictionaries to texts, applying LGs to extract information, generating concordances and comparing concordances.

\begin{figure}[ht]
\begin{center}
\includegraphics[scale=0.6]{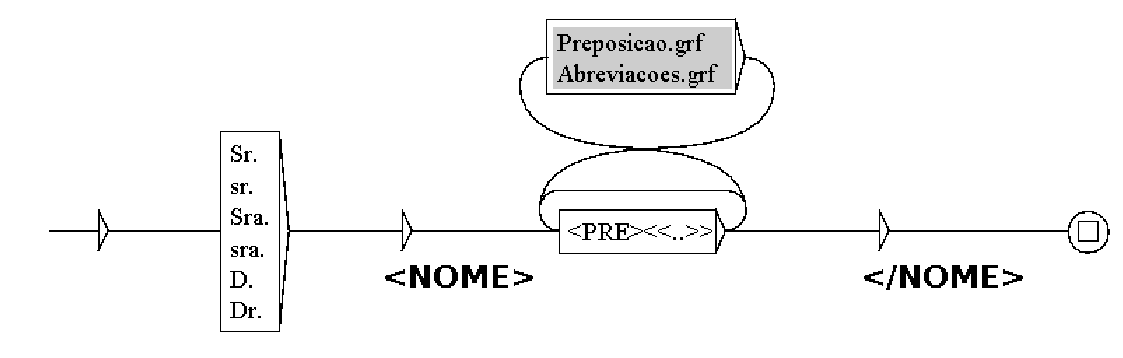}
\end{center}
\caption{LGG {\GI} (ReconheceFormasDeTratamento.grf)}
\label{formasDeTratamento}
\end{figure}

The LGG shown in Fig. \ref{formasDeTratamento} recognizes honorific titles such as \textit{Sr.},\textit{ Sra.} and \textit{Dr.} (``Mr.'', ``Mrs.'', ``Dr.'') followed by
words with the first letter capitalized, as identified by the code \texttt{<PRE>} in Unitex dictionaries. The \texttt{<<..>>} after \texttt{<PRE>} denotes the 
application of a morphological filter to words with the first letter capitalized, indicating that they must include at least two characters. This prevents
the recognition of definite articles at the beginning of sentences, for example. Between the capitalized words, prepositions or abbreviations may occur and are
recognized by two graphs, \textit{Preposicao.grf} and \textit{Abreviacoes.grf}, which have been created separately and are included as subgraphs. Examples of phrases recognized by the graph
(occurrences) include \textit{Sra. Joana da Silva} and \textit{Dr. Ant\^onio de Oliveira Salazar}. A list of occurrences accompanied with one line of context is referred to as a concordance.

Unitex allows for attaching outputs to graph boxes. Outputs are displayed in bold under boxes. In Fig. \ref{formasDeTratamento}, \texttt{<NOME>} (``name'') 
and \texttt{</NOME>} shown under the arrows represent such outputs. Unitex inserts them into the concordance when a graph is applied in the 
``MERGE with input text'' mode. Thus, the identified names appear enclosed in these XML tags in the concordance file. 

The LGs of the repository are small but can be combined to compose a larger grammar to identify person names.

We applied the LGs of the repository to the Golden Collection (GC) of the Second HAREM, producing a concordance
file for each LG. We used Portuguese and English dictionaries because several English names appear in GC texts.

The GC of the Second HAREM \cite{mota2008} is a subset of 129
annotated texts. These texts have different textual genres and are written 
in European or Brazilian Portuguese. The HAREM classifies ten categories of NEs: abstraction, event, thing, place, work, organization, person, time, value, and other.
Person names, the focus of this work, are classified as a subtype within the `person' category
and are represented by the code PERSON (INDIVIDUAL). In the GC of the Second HAREM, 1,609 NEs are 
annotated with this code.

\subsection{Concordance comparison}

We compared all the concordances pairwise (every pair of files) using the ConcorDiff concordance comparison tool
provided by Unitex. This tool can be applied to any pair of concordance files, provided they are in the Unitex format, which is publicly documented in the
manual \cite{paumier2016}.

\begin{figure}[h]
\begin{center}
\includegraphics[scale=0.5]{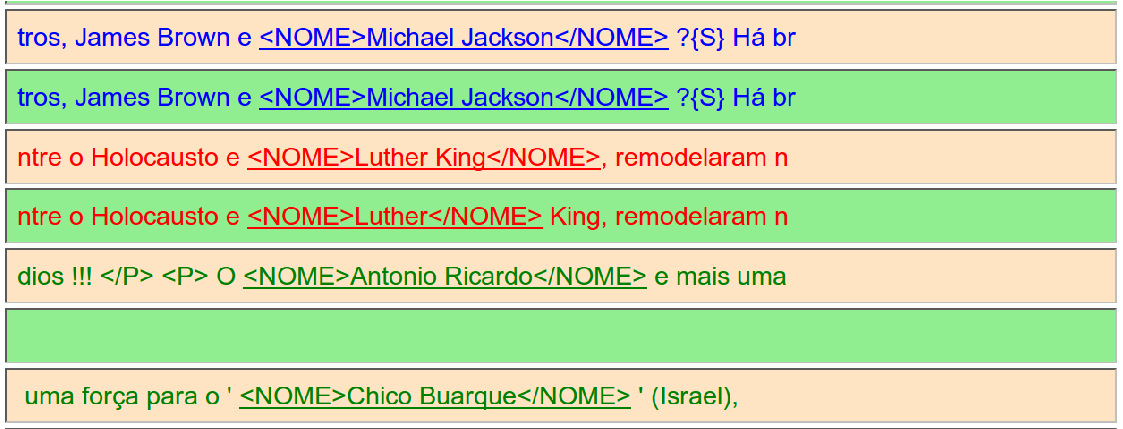}
\end{center}
\caption{Part of a concordance comparison file}
\label{concorDiff}
\end{figure}

The Unitex ConcorDiff program compares two concordance files 
line by line and shows their differences. The result is an HTML page that presents 
alternate lines of both concordances and that leaves an empty line when an occurrence appears
in only one of them. An example is presented in Fig. \ref{concorDiff}. The lines with a pink background 
shading (lines 1, 3, 5 and 7) are from the first concordance (the first parameter
to ConcorDiff), and those with a green background shading (lines 2, 4 and 6) are from
the other concordance (the second parameter to ConcorDiff).

Lines in blue characters (lines 1 and 2) are the occurrences common to
the two concordances. In the example shown in Fig. \ref{concorDiff}, this
means that both LGs 
recognized \textit{Michael Jackson}. Lines in red characters (lines 3 and 4)
correspond to occurrences that overlap only partially, which is the case,
for instance, when an occurrence in a concordance is part of an occurrence in the other. In
the example, an LG recognized \textit{Luther King}, and the
other recognized \textit{Luther}. Lines in  green characters (lines 5 and 7)
are the occurrences that appear in only one of the two
concordances. \textit{Antonio Ricardo} and \textit{Chico Buarque} were
recognized only by the first LG. Lines in purple characters
indicate identical occurrences with different
outputs inserted, which does not happen in this example.

We then analyzed the files generated by ConcorDiff.

\input{methodology-relations.tex}

%% file: methodology-relations.tex
%
%

\subsection{Composition of LG from concordance comparisons}

Let $G_X$ and $G_Y$ be two LGs, and let $C_X$ and $C_Y$ the respective concordance
files obtained by applying them to the same corpus. Thus, $C_X$ is the set of occurrences identified by $G_X$, and $C_Y$ is the set of
occurrences identified by $G_Y$. Let $C_X \times C_Y$ be the file that shows the differences between concordances $C_X$ and $C_Y$ and is obtained
through the ConcorDiff program of Unitex. In $C_X \times C_Y$, the elements $x_1$, $x_2$, ..., $x_n$ of $C_X$ are displayed on a pink background, while the elements
$y_1$, $y_2$, ..., $y_m$ of $C_Y$ are displayed on a green background.  It may exist between $C_X$ and $C_Y$ some relationships of the set theory, such as inclusion, intersection or disjunction, and these relationships can be observed by analyzing $C_X \times C_Y$.

\begin{figure}[ht]
\begin{center}
\includegraphics[scale=0.5]{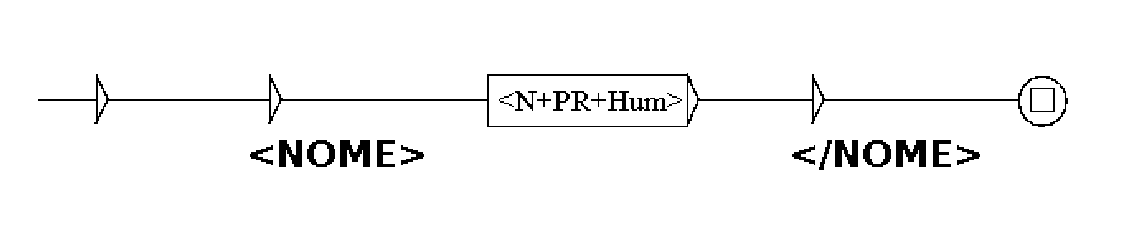}
\end{center}
\caption{LG {\GII} (ReconheceNomesCompostos.grf)}
\label{nomesCompostos}
\end{figure} 

Consider, for example, LGs {\GI} (Fig. \ref{formasDeTratamento}) and {\GII} (Fig. \ref{nomesCompostos}).
{\GII} recognizes person names stored in dictionaries, through dictionary codes \texttt{N+PR} for proper names and \texttt{Hum} for nouns referring to human beings. Multiword person 
names such as \textit{Marilyn Monroe, Cameron Diaz} and \textit{Albert Einstein} are recognized by this LG after applying the English dictionary to the input text.

\begin{figure}[ht]
\begin{center}
\includegraphics[scale=0.45]{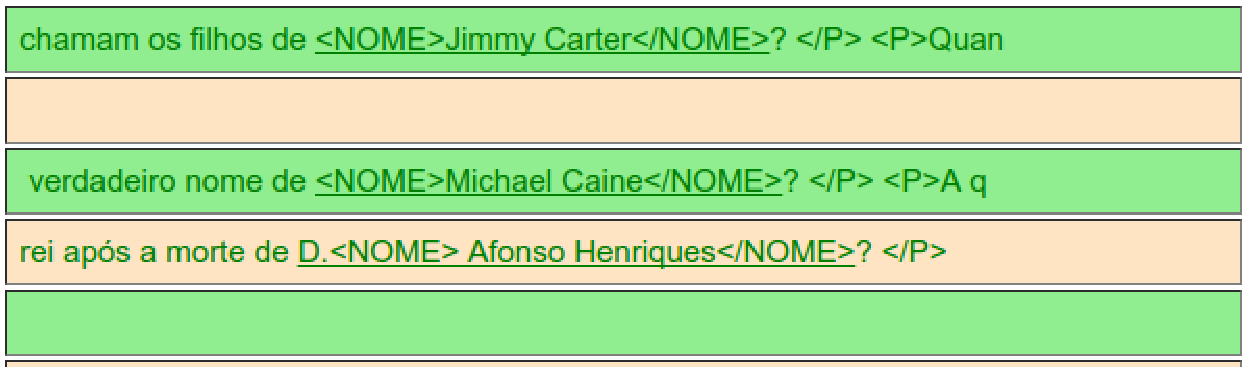}
\end{center}
\caption{Part of the concordance comparison \CIxCII}
\label{concorDiffC1xC2}
\end{figure} 

Figure \ref{concorDiffC1xC2} shows part of the concordance comparison \CIxCII. The first line, $y_1$, includes the name \textit{Jimmy Carter} recognized by \GII. The first line displayed on a pink background, $x_1$, 
 includes the name \textit{Afonso Henriques} occurring after \textit{D.} and recognized by \GI.
Since lines in green characters are occurrences identified by only one of the two graphs, the first two occurrences were identified by {\GII} only,
and the last one by {\GI} only. If all the lines of the comparison are in green characters and distributed between the two background colors,  {\CI} and {\CII} are disjoint sets: thus, both LGs {\GI} and
{\GII} are worth retaining as subgraphs of a grammar because they recognize different names. 

Table \ref{relacoesGramaticas} summarizes the main set-theoretic relationships identified.
Each situation has a consequence in terms of priority between LGs, for example: $G_X$ can be discarded if $G_Y$ is retained.
After analysing relationships between all pairs of LGs, we selected a subset of LGs and combined them into a larger LG (30 LGGs)
by invoking them in a main graph.

\begin{table}[h] \footnotesize
\begin{center}
\caption{Main relationships observed through concordance comparison}
\begin{threeparttable}[b]
\begin{tabular}{p{2cm}|p{3,5cm}|p{3,5cm}|p{3,5cm}}
\hline
\textbf{Relation}                  					& \multicolumn{1}{c}{\textbf{Situation}} & \multicolumn{1}{c}{\textbf{Character color}} & \multicolumn{1}{c}{\textbf{Consequence}}                            	\\ \hline
\multirow{4}{*}{\textbf{Inclusion}}          	& $C_X$ $\subset$ $C_Y$ 														& Blue and green (on green background)        	& Keep $G_Y$   			\\ \cline{2-4}
											& $C_Y$ $\subset$ $C_X$    														& Blue and green (on pink background)         	& Keep $G_X$   			\\ \hline
\multirow{5}{*}{\textbf{Intersection}}		& $C_X$ = $C_Y$                                     											& Blue                                          			& Keep or $G_X$ or $G_Y$ 	\\ \cline{2-4} 
											& $C_X$ = $C_Y$ with different outputs          										& Violet                                        			& Analyze ambiguity		\\ \cline{2-4} 
											& $C_X$ $\cap$ $C_Y$ $\ne \emptyset$                									& Blue and green (on different backgrounds)	& Keep $G_X$ and $G_Y$        	\\ \hline
\multirow{4}{\linewidth}{\textbf{Disjunction}}	& $C_X$ $\cap$ $C_Y$  =  $\emptyset$, with $C_X$ = $\emptyset$ 			& Green (on green background)               		& Keep $G_Y$             	\\ \cline{2-4} 
											& $C_X$ $\cap$ $C_Y$  =  $\emptyset$, with $C_Y$ = $\emptyset$ 				& Green (on pink background)                 		& Keep $G_X$             	\\ \cline{2-4} 
											& $C_X$ $\cap$ $C_Y$  =  $\emptyset$                									& Green (on different backgrounds)           		& Keep $G_X$ and $G_Y$          \\ \hline
\multirow{6}{\linewidth}{\textbf{Disjunction with partial overlapping of occurrences }} & $C_X$ $\cap$ $C_Y$  =  $\emptyset$, with $C_X$ $\sim$ $C_Y$\tnote{1} 	& Red    	& Keep $G_X$ if $\forall i \;  |x_{i}|>|y_{i}|$, keep $G_Y$ if $\forall i \;  |x_{i}|<|y_{i}|$  \\ \cline{2-4} 
											& $C_X$ $\cap$ $C_Y$  =  $\emptyset$, with $\exists i \; \exists j \;\; x_{i}$ overlaps $y_{j}$ 	& Red and green (on identical background)      	& Keep $G_X$ and $G_Y$  if the occurrences in green characters are relevant. If not, keep only the LG that matches larger occurrences                     \\ \hline
\end{tabular}\begin{tablenotes}
  \item[1]{{\footnotesize{$C_X\sim C_Y \Leftrightarrow (n=m$ and $\forall i \; \; x_{i}$ overlaps $y_{i})$. }}}
\end{tablenotes}
\end{threeparttable}
\label{relacoesGramaticas}
\end{center}
\end{table}

%% file: results-en.tex
%
%

We could not compare the performance of the obtained LG to the initial set of small LGs, since this set does not make up a single annotator together. Instead, we simply
evaluated two annotators, one based on the obtained LG and another on an enhanced version of it, and we compared the results to those of Rembrandt, as a widely known 
reference.

We applied the obtained LG to the HAREM corpus and generated an XML file with the identified NEs, annotated according to directives of the Second HAREM. 
Parts of the person names identified by LG that appear isolated in the text are also annotated.

This file was submitted to SAHARA \cite{sahara2018:site} for performance evaluation. SAHARA is an online system for automatic evaluation for HAREM,
which computes the precision, recall and F-measure of an NER system after the user configures the evaluation and submits XML-annotated files.

The results obtained by applying the LG to the GC of the Second HAREM were 59.06\% for precision, 55.22\% for recall and 57.07 for F-measure.

Then, we employed manual strategies to improve the performance of the LG. In the Second HAREM, some words in lowercase letters should 
form part of NE\footnote{http://www.linguateca.pt/aval\_conjunta/HAREM/minusculas.html}. For example, the honorific titles recognized by LGG in
Fig. \ref{formasDeTratamento} and the person's social position that appears before the name. In an example provided by HAREM,\footnote{http://www.linguateca.pt/aval\_conjunta/HAREM/ExemplarioSegundoHAREM.pdf}
\textit{A rainha Isabel II surpreendeu a Inglaterra} ``Queen Elizabeth II surprised England'', not only the name \textit{Isabel}, but the whole phrase \textit{rainha Isabel~II} ``Queen Elizabeth II'' should be labeled as a person name.

We adapted the LGG ReconheceFormasDeTratamento.grf to address this issue by simply shifting the tag (\texttt{<NOME>})  before the honorific title in the graph, so
that the title belongs to the tagged NE. Furthermore, we also used these words in lowercase
letters to recognize the `position' subcategory of the `person' category, represented by PERSON(POSITION), and to recognize person names with a noun of social position in the left context.

The results obtained by the final LG are presented in Table \ref{RembrandtxLG_PN}. They were obtained
with SAHARA by selecting the custom setting PERSON(INDIVIDUAL). This table also shows measures computed by SAHARA for Rembrandt, the system
with the best performance for the `person' category of the Second HAREM. 

\begin{table}[h]
\centering
\caption{Results considering PERSON(INDIVIDUAL): Rembrandt vs. final LG}
\label{RembrandtxLG_PN}
\begin{tabular}{lllll}
\cline{1-4}
\multicolumn{1}{l}{\textbf{System}}   & \multicolumn{1}{l}{\textbf{Precision (\%)}}  & \multicolumn{1}{l}{\textbf{Recall (\%)}}      & \multicolumn{1}{l}{\textbf{F-Measure (\%)}} &  \\ \cline{1-4}
\multicolumn{1}{l}{Rembrandt} 	       & \multicolumn{1}{l}{79}               	      & \multicolumn{1}{l}{64.08}              	      & \multicolumn{1}{l}{70.76}            &  \\ 
\multicolumn{1}{l}{{\thisApproach}}    & \multicolumn{1}{l}{\Precision}               & \multicolumn{1}{l}{\Recall}                   & \multicolumn{1}{l}{\Fmeasure}         &  \\ \cline{1-4}
\end{tabular}
\end{table}

The LG outperfoms Rembrandt. The recall of the LG is approximately 10 percentage points
above that of Rembrandt.

Although our LG recognizes only the `individual' and `position' subtypes of the `person' category, 
its evaluation was also carried out using SAHARA for all types of categories by 
selecting the PERSON(*) setting. A comparison of the obtained results with the results
of the four tools presented in \cite{amaral2014:lrec} for the `person' category  is shown in Table \ref{SystemsxLG_PN}. 

\begin{table}[h]
\centering
\caption{Results considering PERSON(*): Systems in \protect\cite{amaral2014:lrec} vs. final LG}
\label{SystemsxLG_PN}
\begin{tabular}{lllll}
\cline{1-4}
\multicolumn{1}{l}{\textbf{Systems}}          & \multicolumn{1}{l}{\textbf{Precision (\%)}} & \multicolumn{1}{l}{\textbf{Recall (\%)}} & \multicolumn{1}{l}{\textbf{F-Measure (\%)}} &  \\ \cline{1-4}
\multicolumn{1}{l}{NERP-CRF}       	      & \multicolumn{1}{l}{57}               & \multicolumn{1}{l}{51}            & \multicolumn{1}{l}{54}                   &  \\ 
\multicolumn{1}{l}{Freeling}       	      & \multicolumn{1}{l}{55}               & \multicolumn{1}{l}{61}            & \multicolumn{1}{l}{58}                   &  \\ 
\multicolumn{1}{l}{Language-Tasks} 	      & \multicolumn{1}{l}{63}               & \multicolumn{1}{l}{62}            & \multicolumn{1}{l}{62}                   &  \\ 
\multicolumn{1}{l}{PALAVRAS}       	      & \multicolumn{1}{l}{61}               & \multicolumn{1}{l}{65}            & \multicolumn{1}{l}{63}                   &  \\ 
\hline
\multicolumn{1}{l}{{\thisApproach}}           & \multicolumn{1}{l}{\PrecisionPG}     & \multicolumn{1}{l}{\RecallPG}     & \multicolumn{1}{l}{\FmeasurePG}      &  \\ \cline{1-4}                           
\end{tabular}
\end{table}

The LG has a better precision. However, as expected, it has a lower recall as it identifies fewer types of NEs:
only two subtypes of the `person' category (`individual' and `position') are recognized, whereas the other 
systems recognize eight subtypes. We believe that with the addition of rules to the LG in order to recognize 
other subtypes of the `person' category, the recall could be further increased, improving the LG approach 
even more as compared to other tools.

%% file: conclusion-en.tex
%
%

This paper presented the use of the Unitex concordance comparison tool as a computational aid in 
manual composition of LGs. We used this tool for the composition of an LG to identify
person names in texts written in Portuguese. The same methodology can be applied to the
construction of LGs for other purposes.

Table \ref{relacoesGramaticas} was created by listing the main set-theoretic relationships
(inclusion, intersection and disjunction) that we could observe when analyzing 
concordance-comparison files generated by Unitex. 
Taking into account these relationships, we could produce a more compact and easily understandable grammar.
We could also observe that a concordance  
offers an overview of what a LG recognizes in a specific corpus, allowing ambiguities and false positives to be identified.

The results of out final LG show its potential for NE extraction. It performed better (gain of 6 points)
than Rembrandt, the system with the best performance for the `person' category
in the Second HAREM, when evaluating the `person' category, `individual' subtype, for which it was created. 

As avenues for future work, we plan to apply the LG approach to other corpora of
texts written in Portuguese, and to assess performance with a corpus not used in the construction of the LG. 
Moreover, we may add rules for recognizing other types of NEs. We also intend to study the 
feasibility of building elementary LGGs automatically or semi-automatically from examples, with
the goal of minimizing human effort during construction. The concordance comparison tool
presented in this article might facilitate the automation of decision-making for this purpose.